# A new approach to updating beliefs

Ronald Fagin
Joseph Y. Halpern
IBM Almaden Research Center
San Jose, CA 95120

email: fagin@ibm.com, halpern@ibm.com

**Abstract:** We define a new notion of *conditional belief*, which plays the same role for Dempster-Shafer belief functions as conditional probability does for probability functions. Our definition is different from the standard definition given by Dempster, and avoids many of the well-known problems of that definition. Just as the conditional probability $Pr(\cdot|B)$ is a probability function which is the result of conditioning on $B$ being true, so too our conditional belief function $Bel(\cdot|B)$ is a belief function which is the result of conditioning on $B$ being true. We define the conditional belief as the *lower envelope* (that is, the infimum) of a family of conditional probability functions, and provide a closed-form expression for it. We show by example the intuitive appeal of our definition, and compare it in detail to the more standard definition, showing why and how it differs.

## 1 Introduction

How should one update one's belief given new evidence? If beliefs are expressed in terms of probability, then the standard approach is to use conditioning. If an agent's original estimate of the probability of $A$ is given by $Pr(A)$, and then some new evidence, say $B$, is acquired, then the new estimate is given by the conditional probability $Pr(A|B)$, defined as $Pr(A \cap B)/Pr(B)$.[1]

The Dempster-Shafer approach to reasoning about uncertainty [Sha76] has recently become quite popular in expert systems applications (see, for example, [Abe88, Fal88, LU88, LG83]). This approach uses *belief functions*, a class of functions that satisfy three axioms, somewhat related to the axioms of probability. In this paper, we consider how to define a notion of *conditional belief*, which generalizes conditional probability.

One definition for conditional belief was already suggested by Dempster [Dem67], and is derived using the *rule of combination*; hereafter we refer to Dempster's definition as *the DS definition of conditional belief*. Although the DS definition also generalizes conditional probability, it is well known to give counterintuitive results in a number of situations (see, e.g., [Ait68, Bla87, Dia78, DZ82, Hun87, Lem86, Pea88, Pea89, Zad84]) We provide here a new definition of conditional belief, which also generalizes conditional probability, but is different from the DS definition in general. We can show that our definition avoids many of the problems associated with the DS definition.

The motivation for our definition of conditional belief comes from probability theory. It is well known that a belief function $Bel$ is the *lower envelope* of the family of all probability functions $Pr$ consistent with $Bel$. That is, $Bel(A)$ is the infimum of $Pr(A)$, where the infimum is taken over all probability functions $Pr$ such that $Bel(A') \leq Pr(A')$ for all $A'$.[2] We define $Bel(A|B)$ to be the lower envelope of the family of all functions $Pr(\cdot|B)$ where $Pr$ is consistent with $Bel$ (similarly to the situation with conditional probability, we assume that $Bel(B) > 0$, so that everything is well defined). Although we define $Bel(\cdot|B)$ in terms of a lower envelope, we show that there is an elegant closed form expression for it. Moreover, we can show that just as the conditional probability function is in fact a probability function, our conditional belief function is a belief function.

The rest of this paper is organized as follows. In the next section, we review belief functions and define our notion conditional belief. We show how it compares to the DS notion by applying both defi-nitions to the well-known *three prisoners problem* [Gar61, Dia78]. We then conduct a more thorough investigation of the differences between the two notions, and their relationship to conditional probability, showing why the DS notion occasionally provides counterintuitive answers. In Section 4 we discuss the relationship between belief functions and sets of probability functions. We conclude in Section 5 with some discussion on the implications of our results to the use of belief functions.

---

[1] This definition is not completely uncontroversial (see, e.g., [DZ82] for a discussion and further references).

[2] Some authors (e.g., [DP88]) have used the term *lower probability* to denote what we are calling lower envelopes. We have used the term lower envelope here to avoid confusion with Dempster's technical usage of the phrase lower probability in [Dem67, Dem68], which, although related, is not equivalent to what we are calling a lower envelope.

## 2  Updating belief functions

Recall that a probability space is a tuple $(S, \mathcal{X}, Pr)$, where $S$ is the *sample space*, $\mathcal{X}$ is a collection of subsets of $S$ containing $S$ and closed under complementation and countable union, and $Pr$ is a probability function defined on $\mathcal{X}$. Note that $Pr$ is defined not on all subsets of $S$, but only on the sets in $\mathcal{X}$, traditionally called the *measurable* sets. Subsets of $S$ not in $\mathcal{X}$ are called *nonmeasurable*.

The Dempster-Shafer theory of evidence [Sha76] provides an approach to attaching likelihoods to events that is different from probability theory. The theory starts out with a *belief function*. For every event (i.e., set) $A$, the belief in $A$, denoted $Bel(A)$, is a number in the interval $[0, 1]$ that places a lower bound on likelihood of $A$. We have a corresponding number $Pl(A) = 1 - Bel(\overline{A})$, called the *plausibility* of $A$, which places an upper bound on the likelihood of $A$. Thus, to every event $A$ we can attach the interval $[Bel(A), Pl(A)]$. Like a probability measure, a belief function assigns a "weight" to subsets of a set $S$, but unlike a probability measure, the domain of a belief function is always taken to be *all* subsets of $S$. Formally, a belief function $Bel$ on a set $S$ is a function $Bel: 2^S \to [0, 1]$ satisfying:

**B0.** $Bel(\emptyset) = 0$

**B1.** $Bel(A) \geq 0$

**B2.** $Bel(S) = 1$

**B3.** $Bel(A_1 \cup \ldots \cup A_k) \geq$
$\sum_{I \subseteq \{1,\ldots,k\}, I \neq \emptyset} (-1)^{|I|+1} Bel(\bigcap_{i \in I} A_i)$.

We remark that a probability function defined on all of $2^S$ is easily seen to be a belief function.

In a companion paper [HF90], we argue that there are two quite distinct ways of relating belief functions to probability theory. One approach views belief as a generalized probability; the second views it as a way of representing evidence. If we would like to update beliefs, then it seems most appropriate to view beliefs as generalized probabilities. There are a number of ways of doing this [Dem67, Dem68, FH89b, Kyb87, Sha79]. We focus on one here, since it is perhaps the most well known: that is the approach of viewing a belief function as an infimum of a family of probability functions. Given a set $\mathcal{P}$ of probability functions all defined on a sample space $S$, define the *lower envelope* of $\mathcal{P}$ to be the function $f$ such that for each $A \subseteq S$, we have $f(A) = \inf\{Pr(A) : Pr \in \mathcal{P}\}$. We have the corresponding definition of the *upper envelope* of $\mathcal{P}$. It was already known to Dempster [Dem67] that a belief function can be viewed as a lower envelope. More formally, let $Bel$ be a belief function defined on $S$, and let $(S, \mathcal{X}, Pr)$ be a probability space with sample space $S$. We say that $Pr$ is *consistent with* $Bel$ if $Bel(A) \leq Pr(A) \leq Pl(A)$ for each $A \in \mathcal{X}$. Intuitively, $Pr$ is consistent with $Bel$ if the probabilities assigned by $Pr$ are consistent with the intervals $[Bel(A), Pl(A)]$ given by the belief function $Bel$. It is easy to see that $Pr$ is consistent with $Bel$ if $Bel(A) \leq Pr(A)$ for each $A \in \mathcal{X}$ (that is, it follows automatically that $Pr(A) \leq Pl(A)$ for each $A \in \mathcal{X}$). This is because $Pl(A) = 1 - Bel(\overline{A}) \geq 1 - Pr(\overline{A}) = Pr(A)$. Let $\mathcal{P}_{Bel}$ be the set of all probability functions defined on $2^S$ consistent with $Bel$. The next theorem tells us that the belief function $Bel$ is the lower envelope of $\mathcal{P}_{Bel}$, and $Pl$ is the upper envelope.

**Theorem 2.1:** *Let $Bel$ be a belief function on $S$. Then for all $A \subseteq S$,*

$$Bel(A) = \inf_{Pr \in \mathcal{P}_{Bel}} Pr(A)$$
$$Pl(A) = \sup_{Pr \in \mathcal{P}_{Bel}} Pr(A).$$

We remark that the converse to Theorem 2.1 does not hold: not every lower envelope is a belief function. Counterexamples are well known [Bla87, Dem67, Kyb87]. We return to this issue in Section 3.

Theorem 2.1 suggests how we might update a belief function to a *conditional belief function* (and a plausibility function to a *conditional plausibility function*). We define:

$$Bel(A|B) = \inf_{Pr \in \mathcal{P}_{Bel}} Pr(A|B)$$
$$Pl(A|B) = \sup_{Pr \in \mathcal{P}_{Bel}} Pr(A|B).$$

It is not hard to see that the infimum and supremum above are not well-defined unless $Bel(B) > 0$; therefore, we define $Bel(A|B)$ and $Pl(A|B)$ only if $Bel(B) > 0$. It is straightforward to check that if $Bel$ is actually the probability function $Pr$, then $Bel(A|B) = Pr(A|B)$. Thus, our definition of conditional belief generalizes that of conditional probability. Note that taking $B = true$ in the preceding

definition, we get as a special case that $Bel(A) = \inf_{Pr \in \mathcal{P}_{Bel}} Pr(A)$ and $Pl(A) = \sup_{Pr \in \mathcal{P}_{Bel}} Pr(A)$, which is Theorem 2.1 above.

Our definition of conditional belief and plausibility does not give us much help in computing these expressions. We would like to have a closed-form expression for them. We can in fact provide an elegant closed-form expression, as shown in the following theorem.

**Theorem 2.2:** *If Bel is a belief function on S such that $Bel(B) > 0$, then*

$$Bel(A|B) = \frac{Bel(A \cap B)}{Bel(A \cap B) + Pl(\overline{A} \cap B)}$$

$$Pl(A|B) = \frac{Pl(A \cap B)}{Pl(A \cap B) + Bel(\overline{A} \cap B)}.$$

The expressions given above for conditional belief and plausibility are quite natural. Not surprisingly, it turns out that other authors have discovered them as well. In particular, essentially these expressions appear in [Wal81], [SK89], and [dCLM90]. Indeed, it even appears (lost in a welter of notation) as Equation 4.8 in [Dem67]! (Interestingly, none of these papers references any other work as the source of the formula.)

It is well known that the conditional probability function is a probability function. That is, if we start with a probability function $Pr$ on $S$, and $B$ is a subset of $S$ such that $Pr(B) > 0$, then the function $Pr(\cdot|B)$ is a probability function. We might hope that the same situation holds with belief functions, so that the conditional belief and plausibility functions are indeed belief and plausibility functions. Given our definitions of conditional belief and plausibility as lower and upper envelopes, it is not clear that this should be so, since lower and upper envelopes of arbitrary sets of probability functions do not in general result in belief and plausibility functions. Fortunately, as the next result shows, in this case they do. Thus, we have a way of updating belief and plausibility functions to give us new belief and plausibility functions in the light of new information.

**Theorem 2.3:** *Let Bel be a belief function defined on S, and Pl the corresponding plausibility function. Let $B \subseteq S$ be such that $Bel(B) > 0$. Then $Bel(\cdot|B)$ is a belief function and $Pl(\cdot|B)$ is the corresponding plausibility function.*

The proof of Theorem 2.3 is somewhat difficult. We outline the proof in the appendix; full details can be found in the full paper [FH89a]. We remark that this result—which we view as the main technical result of the paper—appears in none of the papers cited above that contain the expression for conditional belief from Theorem 2.2. In [dCLM90] the question of whether $Bel(\cdot|B)$ is a belief function is discussed, but left unanswered.

As we mentioned in the introduction, our definition is quite different from that given by Dempster. Given a belief function $Bel$, Dempster defines a conditional belief function $Bel(\cdot||B)$ as follows [Sha76, p. 97]:[3]

$$Bel(A||B) = \frac{Bel(A \cup \overline{B}) - Bel(\overline{B})}{1 - Bel(\overline{B})}.$$

The corresponding plausibility function is shown to satisfy:

$$Pl(A||B) = \frac{Pl(A \cap B)}{Pl(B)}.$$

A brief glance at the DS definition compared with the formula in Theorem 2.2 should convince the reader that in general these two definitions of conditional belief will not agree. It is easy to show that both definitions of conditional belief generalize the standard definition of conditional probability as long as all sets are *measurable*, that is, have a probability assigned to them. The key difference turns out to be in the way they treat nonmeasurable sets. (See [FH89b] for a discussion of nonmeasurable sets and their relationship to belief functions.) We return to this issue below; we first consider an example that highlights the differences between the two approaches.

**Example:** In order to compare our updating technique with that of Dempster, we consider the well-known *three prisoners problem*.[4]

> Of three prisoners $a$, $b$, and c, two are to be executed but $a$ does not know which. He therefore says to the jailer, "Since either $b$ or $c$ is certainly going to be executed, you will give me no information about my own chances if you give me the name of one man, either $b$ or $c$, who is going to be executed." Accepting this argument, the jailer truthfully replies, "$b$ will be executed." Thereupon $a$ feels happier because before the jailer replied, his own chance of execution was two-thirds, but afterwards there are only two people, himself and $c$, who could be the one not executed, and so his chance of execution is one-half.

---

[3] Dempster's definition is usually given as a special application of a more general *rule of combination* for belief functions. It would take us too far afield here to discuss the rule of combination; see the companion paper [HF90] for a discussion of the role of the rule of combination.

[4] For an excellent introduction to the problem as well as a Bayesian solution, see [Gar61]. Our description of the story is taken from [Dia78] and much of our discussion is based on that of Diaconis and Zabell [Dia78, DZ82].

Note that in order for $a$ to believe that his own chance of execution was two-thirds before the jailer replied, he seems to be implicitly assuming that the one to get pardoned is chosen at random from among $a$, $b$, and $c$. We make this assumption explicit in the remainder of our discussion.

Is $a$ justified in believing that his chances of escaping have improved? It seems that the jailer did not give him any relevant extra information. Yet how could $a$'s subjective probabilities change if he does not acquire any relevant extra information?

Following [DZ82], we model a possible situation by an ordered pair $(x, y)$, where $x, y \in \{a, b, c\}$. Intuitively, a pair $(x, y)$ represents a situation where $x$ is pardoned and the jailer says that $y$ will be executed in response to $a$'s question. Since the jailer answers truthfully, we cannot have $x = y$; since the jailer will never tell $a$ directly that $a$ will be executed, we cannot have $y = a$. Thus, the set of possible outcomes is $\{(a, b), (a, c), (b, c), (c, b)\}$. The event that $a$ lives, which we denote *lives-a*, corresponds to the set $\{(a, b), (a, c)\}$. Similarly, we define the events *lives-b* and *lives-c*, which correspond to the sets $\{(b, c)\}$ and $\{(c, b)\}$, respectively. By assumption, each of these three events has probability $1/3$.

The event that the jailer says $b$, which we denote *says-b*, corresponds to the set $\{(a, b), (c, b)\}$; the story does not give us a probability for this event. In order to do a Bayesian analysis of the situation, we will need this probability. Note that we do know that the probability of $\{(c, b)\}$ is $1/3$; we just need to know the probability of $\{(a, b)\}$. This depends on the jailer's strategy in the one case that he has free choice, namely when $a$ lives. He gets to choose between saying $b$ and $c$ in that case. We need to know the probability that he says $b$, i.e., $Pr(\textit{says-b}|\textit{lives-a})$.

If we assume that the jailer chooses at random between saying $b$ and $c$ if $a$ is pardoned, so that $Pr(\textit{says-b}|\textit{lives-a}) = 1/2$, then $Pr(\{(a, b)\}) = Pr(\{(a, c)\}) = 1/6$, and $Pr(\textit{says-b}) = 1/2$. We can now easily compute that

$$Pr(\textit{lives-a}|\textit{says-b})$$
$$= Pr(\textit{lives-a} \cap \textit{says-b})/Pr(\textit{says-b})$$
$$= (1/6)/(1/2) = 1/3.$$

Thus, in this case, the jailer's answer does not affect $a$'s probability.

Suppose more generally that $Pr(\textit{says-b}|\textit{lives-a}) = \alpha$, for $0 \leq \alpha \leq 1$. Then straightforward computations show that

$$Pr(\{(a, b)\}) = Pr(\textit{lives-a}) \times Pr(\textit{says-b}|\textit{lives-a})$$
$$= \alpha/3$$

$$Pr(\textit{says-b}) = Pr(\{(a, b)\}) + Pr(\{(c, b)\}) = (\alpha + 1)/3$$

$$Pr(\textit{lives-a}|\textit{says-b}) = \frac{\alpha/3}{(\alpha + 1)/3} = \alpha/(\alpha + 1).$$

This says that if $\alpha \neq 1/2$ (i.e., if the jailer had a particular preference for answering either $b$ or $c$ when $a$ was the one pardoned), then $a$ would learn something from the answer, in that he would change his estimate of the probability that he will be executed. For example, if $\alpha = 0$, then if $a$ is pardoned, the jailer will definitely say $c$. Thus, if the jailer actually says $b$, then $a$ knows that he is definitely not pardoned, i.e., that $Pr(\textit{lives-a}|\textit{says-b}) = 0$. Similarly, if $\alpha = 1$, then $a$ knows that if either he or $c$ is pardoned, then the jailer will say $b$, while if $b$ is pardoned the jailer will say $c$. Given that the jailer says $b$, then from $a$'s point of view the one pardoned is equally likely to be him or $c$; thus, $Pr(\textit{lives-a}|\textit{says-b}) = 1/2$. As $\alpha$ ranges from $0$ to $1$, it is easy to check that $Pr(\textit{lives-a}|\textit{says-b})$ ranges from $0$ to $1/2$.

How can we capture this situation using a belief function? It seems reasonable that if $Bel$ is the belief function and $Pl$ the corresponding plausibility function used to capture the situation, then $Bel$ should agree with probability function where the probability is known, so that, *a priori*, both the belief and plausibility of the events *lives-a*, *lives-b*, and *lives-c* should be $1/3$. All we know about the *a priori* probability of *says-b* is that it lies between $1/3$ and $2/3$: it is at least the probability that $c$ is chosen (since in that case the jailer must say $b$), and it cannot be more than the probability that $b$ is not chosen. Thus, we assume that $Bel$ satisfies $Bel(\textit{say-b}) = 1/3$ and $Pl(\textit{say-b}) = 2/3$. Similarly, we can argue that $Bel(\textit{lives-a} \cap \textit{says-b}) = 0$ while $Pl(\textit{lives-a} \cap \textit{says-b}) = 1/3$. Plugging these numbers into our formulas, it is easy to compute that $Bel(\textit{lives-a}||\textit{says-b}) = Pl(\textit{lives-a}||\textit{says-b}) = 1/2$. Thus, for the DS notion of conditional probability, the range reduces to the single point $1/2$. By way of contrast, it is easy to check that $Bel(\textit{lives-a}|\textit{says-b}) = 0$ while $Pl(\textit{lives-a}|\textit{says-b}) = 1/2$. ∎

This example shows that the two notions of conditioning can give quite different answers. The range $[0, 1/2]$ computed by our notion of conditioning is easy to explain: it is precisely the range determined by letting the probability that the jailer will say $b$ in the one situation that he has a choice between saying $b$ and $c$, namely, when $a$ is the one pardoned, range from 0 to 1. The fact that our definition gives this range is not an accident! It is a direct consequence of our definitions and Theorem 2.2.

The range $[1/2, 1/2]$ determined by the DS notion of conditioning seems much more mysterious. The answer $1/2$ corresponds to the situation where the jailer says $b$ whenever he can (i.e., whenever $a$ is pardoned or $c$ is pardoned). Why is this a reasonable answer? More importantly, why does it arise? Is there a natural probabilistic interpretation for it? In the full paper, we consider this issue in detail. The following construction, which is a generalization of the "beehive" example in [SK89] (as well as being a formalization of some comments made in [dCLM90]), may help provide a partial explanation.

Suppose a set $S$ is partitioned into (nonempty) disjoints sets $X_1, \ldots, X_k$. An agent chooses $X_i$ with probability $a_i$ (where $a_1 + \cdots + a_k = 1$) and then chooses $x \in X_i$ with some unknown probability. Given subsets $A$ and $B$ of $S$, we want to know what the probability is that the element $x$ chosen is in $A$, and the probability that $x$ is in $A$ given that it is in $B$. If $A = X_i$, then it is clear that the probability that $x \in A$ is $a_i$. However, if $A$ is not one of the $X_i$'s, then all we can compute are upper and lower bounds on the probability.

Let $\mathcal{P}$ be the set of probability functions on $S$ consistent with this situation; namely, $Pr \in \mathcal{P}$ iff $Pr(X_i) = a_i$, for $i = 1, \ldots, k$. Let $Bel$ be the lower envelope of $\mathcal{P}$; it is not hard to show that $Bel$ is a belief function (we do so in the full paper). It seems reasonable to argue that the best lower and upper bounds we can give on the probability that $x \in A$ are $Bel(A)$ and $Pl(A)$. Similarly, the best lower and upper bounds we can give on the probability that $x \in A$ given that $x \in B$ are given by the infimum and supremum of $\{Pr(A|B) : Pr \in \mathcal{P}\}$. These are precisely $Bel(A|B)$ and $Pl(A|B)$.

Now suppose we slightly change the rules of the game. We are told that the probabilistic process that chooses an element in $X_i$ will definitely choose an element in $B$ if possible. This does not affect anything if $X_i \subseteq B$ or if $X_i \subseteq \overline{B}$. However, if $X_i \cap B \neq \emptyset$ and $X_i \cap \overline{B} \neq \emptyset$, then, rather than choosing $X_i$ with probability $a_i$, the probability is now redistributed so that $X_i \cap B$ is chosen with probability $a_i$, while $X_i \cap \overline{B}$ is chosen with probability 0. The probability that used to be spread over all of $X_i$ is now concentrated on $X_i \cap B$. What is the probability that an element of $A$ is chosen given that the element chosen is definitely in $B$ with respect to this new process, where an element of $B$ is chosen whenever possible? We now have to consider the family $\mathcal{P}'$ of probability functions consistent with this new process, and take the infimum and supremum of $\{Pr'(A|B) : Pr \in \mathcal{P}'\}$. As we show in the full paper, these bounds are given by $Bel(A\|B)$ and $Pl(A\|B)$.

Suppose we now reconsider the three prisoners problem from this point of view. We can now see that $Bel(lives\text{-}a\|says\text{-}b)$ gives the probability that $a$ lives given the extra hypothesis that the jailer says $b$ whenever possible. In particular, this means that the jailer definitely says $b$ if $a$ is the one that is pardoned; i.e., $Pr(says\text{-}b|lives\text{-}a) = 1$. Under this revised situation, the probability that $a$ lives given that the jailer says $b$ is indeed exactly $1/2$. With this understanding of the DS notion of updating, the result $Bel(lives\text{-}a\|says\text{-}b) = Pl(lives\text{-}a\|says\text{-}b) = 1/2$ should come as no surprise.

To summarize, this discussion has shown that $Bel(A\|B)$ corresponds to a somewhat unnatural updating process, where before we condition with respect to $B$, we first try to choose an element in $B$ whenever possible. In terms of the process discussed above, it is easy to see that this extra step before updating makes no difference if $B$ is the union of some of the $X_i$'s. This amounts to $B$ being a measurable set. It will make a difference if $B$ is not measurable. This is the case in the three prisoner problem, where $says\text{-}b$ is not a measurable set, and is the cause of the answer $1/2$ that we get when we try to apply DS conditioning in this case.

We remark that this analysis can also be used to explain the well-known observation that $Bel(A|B) \leq Bel(A\|B) \leq Pl(A\|B) \leq Pl(A|B)$ ([Dem67, Dem68]; see also [Kyb87]). Not only does it show *why* the interval defined by $[Bel(A|B), Pl(A|B)]$ contains that defined by $[Bel(A\|B), Pl(A\|B)]$, it explains when and why we get equality. See the full paper for details.

## 3 Belief functions and lower envelopes

Theorem 2.1 says that each belief function is the lower envelope of a set of probability functions, and each plausibility function an upper envelope. Unfortunately, the lower envelope of an arbitrary set of probability functions is not in general a belief function, nor is the upper envelope of an arbitrary set of probability functions in general a plausibility function. Nevertheless, results such as Theorem 2.3 show that there are natural sets of probability functions that do induce belief and plausibility functions. Although a general characterization is lacking, further examples in [FH89b, HF90] suggest that this is not an isolated example.

However, even if a set $\mathcal{P}$ of probability functions does induce a belief and plausibility function, say $Bel$ and $Pl$, it is reasonable to ask whether we *should* represent $\mathcal{P}$ by $Bel$ and $Pl$. Clearly the answer depends very much on the intended application. However, it is worth noting that this representation of $\mathcal{P}$ might result in a loss of valuable information. For example, consider a sample space consisting of three points, say $\{a, b, c\}$. Let $\mathcal{P}$ consist of all probability functions on $S$ with the following three properties: (1) $1/4 \leq Pr(\{a\}) \leq 1/2$, (2) $1/4 \leq Pr(\{b\}) \leq 1/2$, and (3) $Pr(\{a\}) = Pr(\{b\})$. It is not hard to show that the lower envelope of $\mathcal{P}$ is a belief function. If we call this belief function $Bel$ and take $Pl$ to be the corresponding plausibility function, we get $Bel(\{a\}) = Bel(\{b\}) = 1/4$ and $Pl(\{a\}) = Pl(\{b\}) = 1/2$. Thus, we retain the information that the probability of $a$ and $b$ both range between $1/4$ and $1/2$. However, we have lost the information that the probabilities of $a$ and $b$ are the same in all the probability functions in $\mathcal{P}$. This loss of information has some serious repercussions. As we show in the full paper (by extending the example above), one consequence is that updates do not commute. More precisely, suppose we start with a belief function $Bel$ on a set $S$, observe $B \subseteq S$ and then observe $C \subseteq S$. The result is the belief function $Bel_B(\cdot|C)$, where $Bel_B(A) = Bel(A|B)$. Similarly, if we observe $C$ and then $B$, we get the belief function $Bel_C(\cdot|B)$. We might hope that for all sets $A$, we would have $Bel_B(A|C) = Bel_C(A|B) = Bel(A|B \wedge C)$. That is, observing $B$ then $C$ should be the same as observing $C$ then $B$, which in turn should be the same as observing $B \wedge C$. This is certainly the case if $Bel$ is a probability function, but not in general. It turns out that the problem here is that information is lost as we update the belief function. (See the full paper for further details of this issue.) By way of contrast, the DS rule of conditioning is commutative. Conditioning with respect to $C$ and then with respect to $B$ is equivalent to conditioning with respect to $B \wedge C$. However, as we have pointed out, it has other problems when viewed as a technique for updating beliefs.

These observations suggest to us that the question of the "best" representation of evidence does not have a unique answer. It may be easier to compute with a pair of belief and plausibility functions than to have to carry around a whole set of probability functions. Nevertheless, since information may be lost in this process, this ease of computation comes at a cost. (See [Pea89] for further examples of this phenomenon.)

## 4 Conclusions

We have defined a new notion of conditional belief, distinct from the DS notion, that seems to lead to more intuitive results. Our notion also allows us to avoid some paradoxes associated with the DS notion. For example, we would expect that if both an agent's belief in a proposition $p$ given $q$ and his belief in $p$ given $\neg q$ are at least $\alpha$, then his belief in $p$ should be at least $\alpha$, whether or not he learns anything about $q$. This is essentially what Savage [Sav54] has called the *sure thing principle*. It is easy to see that conditional probability satisfies the sure thing principle, but the DS conditioning rule does not (see [Pea89] for an example). On the other hand, it is easy to see that our notion of conditioning does satisfy the sure thing principle. For suppose we have an arbitrary belief function $Bel$ such that $Bel(p|q) \geq \alpha$ and $Bel(p|\neg q) \geq \alpha$. Choose an arbitrary probability function $Pr$ compatible with $Bel$. By our definition of conditional belief as an infimum, we see that $Pr(p|q) \geq \alpha$ and $Pr(p|\neg q) \geq \alpha$. So $Pr(p) \geq \alpha$. Thus, $Pr(p) \geq \alpha$ for all probability functions $Pr$ compatible with $Bel$. So, from Theorem 2.1, it follows that $Bel(p) \geq \alpha$.

Although our results show that belief functions can play a useful role even when one wants to think probabilistically, the observations of the previous section do show that information can be lost if we pass to belief functions. This suggests they should be used with care.

One thing we have not really discussed in this paper is what is considered perhaps the key component of the Dempster-Shafer approach, namely, the *rule of combination*. This rule is a way of combining two belief functions to obtain a third one. The reason we have not discussed it is that we feel that the rule of combination does not fit in well with the viewpoint of belief functions as a generalization of probability functions that is discussed in this paper. However, there is another way of viewing belief functions, which is as representations of evidence. This is in fact the view taken in [Sha76]. When belief is viewed as a representation of evidence, then the rule of combination becomes more appropriate. These issues are discussed in more detail in a companion paper [HF90].

## Appendix: Proof of Theorem 2.3

In order to carry out this proof, it will be technically useful to think in terms of *mass function* rather than belief functions. A *mass function* on $S$ is simply a function $m: 2^S \to [0,1]$ such that

**M1.** $m(\emptyset) = 0$

**M2.** $\sum_{A \subseteq S} m(A) = 1$.

Intuitively, $m(A)$ is the weight of evidence for $A$ that has not already been assigned to some proper subset of $A$. With this interpretation of mass, we would expect that an agent's belief in $A$ is the sum of the masses he has assigned to all the subsets of $A$; i.e., $Bel(A) = \sum_{B \subseteq A} m(B)$. Indeed, this intuition is correct.

**Proposition 4.1:** ([Sha76, p. 39])

1. *If $m$ is a mass function on $S$, then the function $Bel: 2^S \to [0,1]$ defined by $Bel(A) = \sum_{B \subseteq A} m(B)$ is a belief function.*

2. *If $Bel$ is a belief function on $2^S$ and $S$ is finite, then there is a unique mass function $m$ on $2^S$ such that $Bel(A) = \sum_{B \subseteq A} m(B)$ for every subset $A$ of $S$.*

Recall that we want to show $Bel(\cdot|B)$ is a belief function, and $Pl(\cdot|B)$ is the corresponding plausibility function, provided that $Bel(B) > 0$. For simplicity in this proof, we work under the assumption that $S$ is finite, so that there is a mass function $m$ corresponding to $Bel$. We remark that using techniques of [FH89b] we could drop this assumption.

It is easy to see, using the formulas in Theorem 2.2, that

$$\begin{aligned} Pl(\overline{A}|B) &= \frac{Pl(\overline{A} \cap B)}{Pl(\overline{A} \cap B) + Bel(A \cap B)} \\ &= 1 - \frac{Bel(A \cap B)}{Pl(\overline{A} \cap B) + Bel(A \cap B)} \\ &= 1 - Bel(A|B). \end{aligned}$$

Thus, once we show that $Bel(\cdot|B)$ is a belief function, it will immediately follow that $Pl(\cdot|B)$ is the corresponding plausibility function.

Let $Bel'$ be the function defined on $2^B$ such that for each subset $A$ of $B$,

$$Bel'(A) = Bel(A)/(Bel(A) + Pl(\overline{A} \cap B)).$$

It clearly suffices to show that $Bel'$ is a belief function, since for all subsets $C$ of $S$, we have $Bel(C|B) = Bel'(C \cap B)$. Once we show that $Bel'$ satisfies axioms B0–B3, it immediately follows that $Bel(\cdot|B)$ does.

It is clear that $Bel'$ satisfies B0–B2. All that remains is to show that $Bel'$ satisfies B3. Thus we must show that the following inequality holds:

$$\begin{aligned} & Bel'(A_1 \cup \ldots \cup A_k) \\ \geq & \sum_{I \subseteq \{1,\ldots,k\}, I \neq \emptyset} (-1)^{|I|+1} Bel'(\bigcap_{i \in I} A_i). \end{aligned}$$

Let $B_1, \ldots, B_t$ be the distinct sets with positive mass contained in $B$. Let $A'_1, \ldots, A'_n$ be the distinct sets with positive mass that intersect $B$ but are not subsets of $B$, and let $A_i = A'_i \cap B$, for $1 \leq i \leq n$. Since $Bel(B) > 0$, we know that there is some $B_i$ (but there may be no $A_i$). Let $\alpha'_i = m(A'_i)$, and $\beta'_i = m(B_i)$, for each $i$. Let $N = \sum_{i=1}^n \alpha'_i + \sum_{i=1}^t \beta'_i$. Note that $N > 0$, since there is some $B_i$. Let $\alpha_i = \alpha'_i/N$, and $\beta_i = \beta'_i/N$, for each $i$. Thus, the $\alpha_i$'s and $\beta_i$'s are normalized versions of the $\alpha'_i$'s and $\beta'_i$'s.

We want to define a mass function $m'$ corresponding to $Bel'$. We first need to do a small detour. If $s_1 \ldots s_k$ is a string, and if $1 \leq i_1 < \cdots < i_p \leq k$, then we call $s_{i_1} \ldots s_{i_p}$ a *substring of* $s_1 \ldots s_k$, which we write as $s_{i_1} \ldots s_{i_p} \preceq s_1 \ldots s_k$. For example, $s_1 s_3 s_4$ is a substring of $s_1 s_2 s_3 s_4 s_5$. The substring is *proper* if it does not equal the full string $s_1 \ldots s_k$; we then write $s_{i_1} \ldots s_{i_p} \prec s_1 \ldots s_k$. We now define a function $m''$, whose domain is $\{A_1, \ldots, A_n, B_1, \ldots, B_t\}^*$, the set of finite strings over the alphabet consisting of the names of the sets with positive mass that intersect $B$. (We shall usually not bother to distinguish between a set and the name of a set, but, as we shall see, it is convenient to consider explicitly strings of names of sets.) First, we let $m''(B_i) = \beta_i$, for $1 \leq i \leq t$. Assume now that we have defined $m''(B_i A_{j_1} \cdots A_{j_s})$ whenever $s < r$ and $j_1 < \cdots < j_s$. Assume that $j_1 < \cdots < j_r$. Let

$$m''(B_i A_{j_1} \cdots A_{j_r}) = \frac{\beta_i}{1 - \alpha_{j_1} - \cdots - \alpha_{j_r}} \quad (1)$$
$$- \sum_{Y \prec A_{j_1} \cdots A_{j_r}} m''(B_i Y).$$

If $A$ is not of the form $B_i A_{j_1} \cdots A_{j_r}$ with $j_1 < \cdots < j_r$, then $m''(A) = 0$.

We are now ready to define the alleged mass function $m'$. If $X$ is the string $B_i A_{j_1} \cdots A_{j_r}$, where $j_1 < \cdots < j_r$, then we say that $X$ *represents* the set given by $B_i \cup A_{j_1} \cup \cdots \cup A_{j_r}$. We would like to let $m'$ be simply $m''$ (that is, by letting $m'$ applied to a set be equal to $m''$ applied to a string that represents the set, and let $m'(A) = 0$ for sets not of the form $B_i A_{j_1} \cdots A_{j_r}$). The problem is that several distinct strings may represent the same set; for example, it is quite possible that, say, the sets $B_1 \cup A_1$ and $B_2 \cup A_4 \cup A_5$ are the same. We define $m'(A)$ to be $\sum_{X \text{ represents } A} m''(X)$. For example, if the set $A$ equals both $B_1 \cup A_1$ and $B_2 \cup A_4 \cup A_5$, but if $A$ is not of the form $B_i \cup A_{j_1} \cdots \cup A_{j_r}$ for any other choices of $B_i, A_{j_1}, \ldots, A_{j_r}$ with $j_1 < \cdots < j_r$, then $m'(A) = m''(B_1 A_1) + m''(B_2 A_4 A_5)$. We shall prove that $m'$ is a mass function, and that $Bel'(A) = \sum_{C \subseteq A} m'(C)$. This will show that $Bel'$ is a belief function.

Thus, we must show that

**A.** $m'(\emptyset) = 0$.

**B.** $m'(A) \geq 0$, for each $A \subseteq B$.

**C.** $\sum_{A \subseteq B} m'(A) = 1$.

**D.** $Bel'(A) = \sum_{C \subseteq A} m'(C)$.

By definition of $m''$ and $m'$, we know that (A) holds. We now prove (D). Let $A_{k_1}, \ldots, A_{k_q}$ (where $k_1 < \cdots < k_q$) be the $A_i$'s contained in $A$, and let $B_{i_1}, \ldots, B_{i_s}$ be the $B_i$'s contained in $A$. What is $Bel'(A)$? As before, let $N = \sum_{i=1}^{n} \alpha'_i + \sum_{i=1}^{t} \beta'_i$. It is easy to see that $Bel(A) = \beta'_{i_1} + \cdots + \beta'_{i_s}$, and $Pl(\overline{A} \cap B) = N - (\alpha'_{k_1} + \cdots + \alpha'_{k_q} + \beta'_{i_1} + \cdots + \beta'_{i_s})$. Hence,

$$Bel'(A)$$
$$= Bel(A)/(Bel(A) + Pl(\overline{A} \cap B))$$
$$= (\beta'_{i_1} + \cdots + \beta'_{i_s})/(N - \alpha'_{k_1} - \cdots - \alpha'_{k_q}).$$

When we divide numerator and denominator by $N$, we see that

$$Bel'(A) = (\beta_{i_1} + \cdots + \beta_{i_s})/(1 - \alpha_{k_1} - \cdots - \alpha_{k_q}). \quad (2)$$

To prove (D), we must show that $\sum_{C \subseteq A} m'(C)$ equals the right-hand side of (2). Let us call an expression $m''(B_i A_{j_1} \cdots A_{j_r})$, where $i$ is a member of $\{i_1, \ldots, i_s\}$, and where $j_1 < \ldots < j_r$ are members of $\{k_1, \ldots, k_q\}$, a *good term*. Note that if $m''(B_i A_{j_1} \cdots A_{j_r})$ is a good term, then $B_i \cup A_{j_1} \cup \cdots \cup A_{j_r} \subseteq A$. Now $\sum_{C \subseteq A} m'(C)$ equals the sum of all good terms. This is because (a) each good term is a part of the sum defining $m'(C)$ for exactly one $C \subseteq A$, and (b) if $C \subseteq A$, then $m'(C)$ is defined as the sum of certain good terms. So we must show that the sum of all of the good terms equals the right-hand side of (2). Now let $i$ be a fixed member of $\{i_1, \ldots, i_s\}$. The sum of all good terms of the form $m''(B_i A_{j_1} \cdots A_{j_r})$ except for the good term $m''(B_i A_{k_1} \cdots A_{k_q})$ is simply $\sum_{Y \prec A_{k_1} \cdots A_{k_q}} m''(B_i Y)$. Since

$$m''(B_i A_{k_1} \cdots A_{k_q}) = \frac{\beta_i}{1 - \alpha_{k_1} - \cdots - \alpha_{k_q}} - \sum_{Y \prec A_{k_1} \cdots A_{k_q}} m''(B_i Y),$$

it follows that the sum of all good terms of the form $m''(B_i A_{j_1} \cdots A_{j_r})$ equals $\beta_i / (1 - \alpha_{k_1} - \cdots - \alpha_{k_q})$. So the sum of all good terms is $(\beta_{i_1} + \cdots + \beta_{i_s})/(1 - \alpha_{k_1} - \cdots - \alpha_{k_q})$, as desired. This proves (D).

Now (C) follows from (D), since it is easy to see that $Bel'(B) = 1$. So we need only prove (B). The proof of (B) involves some nontrivial combinatorial arguments; the details can be found in the full paper. ∎

We remark that in response to an early draft of this paper, Zhang [Zha89] constructed a proof along very different lines (although also quite complicated).

**Acknowledgments:** The second author would like to thank Judea Pearl for a series of net conversations that inspired the development of the definition of conditional belief. We also gratefully acknowledge Nati Linial for his help in the proof of Theorem 2.3. Comments by Hector Levesque, Philippe Smets, and Moshe Vardi inspired a number of useful changes. Finally, we thank Judea Pearl, Tom Strat, and Sue Andrews, for sending us, respectively, [dCLM90], [SK89], and [Wal81], and Enrique Ruspini for pointing out that our expressions for conditional belief and plausibility actually appear in [Dem67].